\documentclass{article}

\usepackage{times}
\usepackage{amsmath,graphicx}
\usepackage{graphicx} % more modern
\usepackage{subfigure} 
\usepackage{bm}
\usepackage{amssymb}
\usepackage{amsfonts}
\usepackage{color}
\usepackage{multirow}
\usepackage{float}
\usepackage[titletoc,toc]{appendix}

\DeclareMathOperator*{\argmin}{arg\,min}

\usepackage{natbib}

\usepackage{algorithm}
\usepackage{algorithmic}

\usepackage{hyperref}

\usepackage[accepted]{implicitworkshop2017}

\icmltitlerunning{Conditional generation of multi-modal data using constrained embedding space mapping}

\begin{document} 

\twocolumn[
\icmltitle{Conditional generation of multi-modal data \\
			  using constrained embedding space mapping}

% It is OKAY to include author information, even for blind
% submissions: the style file will automatically remove it for you
% unless you've provided the [accepted] option to the icml2017
% package.

% list of affiliations. the first argument should be a (short)
% identifier you will use later to specify author affiliations
% Academic affiliations should list Department, University, City, Region, Country
% Industry affiliations should list Company, City, Region, Country

% you can specify symbols, otherwise they are numbered in order
% ideally, you should not use this facility. affiliations will be numbered
% in order of appearance and this is the preferred way.
\icmlsetsymbol{equal}{*}

\begin{icmlauthorlist}
\icmlauthor{Subhajit Chaudhury}{ibm}
\icmlauthor{Sakyasingha Dasgupta}{ibm}
\icmlauthor{Asim Munawar}{ibm}
\icmlauthor{Md. A. Salam Khan}{ibm}
\icmlauthor{Ryuki Tachibana}{ibm}
%\icmlauthor{Tateu H.~Yasehe}{ed,to,goo} 
%\icmlauthor{Aaoeu Iasoh}{goo}
%\icmlauthor{Buiui Eueu}{ed}
%\icmlauthor{Aeuia Zzzz}{ed}
%\icmlauthor{Bieea C.~Yyyy}{to,goo}
%\icmlauthor{Teoau Xxxx}{ed}
%\icmlauthor{Eee Pppp}{ed}
\end{icmlauthorlist}

\icmlaffiliation{ibm}{IBM Research-Tokyo, Japan}
%\icmlaffiliation{goo}{Googol ShallowMind, New London, Michigan, USA}
%\icmlaffiliation{ed}{University of Edenborrow, Edenborrow, United Kingdom}

\icmlcorrespondingauthor{Subhajit Chaudhury}{subhajit@jp.ibm.com}
%\icmlcorrespondingauthor{Eee Pppp}{ep@eden.co.uk}

% You may provide any keywords that you 
% find helpful for describing your paper; these are used to populate 
% the "keywords" metadata in the PDF but will not be shown in the document
\icmlkeywords{Generative models, Cross-Domain representations, Shared latent space, Language to Image translation}

\vskip 0.3in
]

% this must go after the closing bracket ] following \twocolumn[ ...

% This command actually creates the footnote in the first column
% listing the affiliations and the copyright notice.
% The command takes one argument, which is text to display at the start of the footnote.
% The \icmlEqualContribution command is standard text for equal contribution.
% Remove it (just {}) if you do not need this facility.

\printAffiliationsAndNotice{}  % leave blank if no need to mention equal contribution
%\printAffiliationsAndNotice{\icmlEqualContribution} % otherwise use the standard text.
%\footnotetext{hi}

\begin{abstract}
	We present a conditional generative model that maps low-dimensional embeddings of multiple modalities of data to a common latent space hence extracting semantic relationships between them. The embedding specific to a modality is first extracted and subsequently a constrained optimization procedure is performed to project the two embedding spaces to a common manifold. The individual embeddings are generated back from this common latent space. However, in order to enable independent conditional inference for separately extracting the corresponding embeddings from the common latent space representation, we deploy a proxy variable trick - wherein, the single shared latent space is replaced by the respective separate latent spaces of each modality. We design an objective function, such that, during training we can force these separate spaces to lie close to each other, by minimizing the distance between their probability distribution functions. Experimental results demonstrate that the learned joint model can generalize to learning concepts of double MNIST digits with additional attributes of colors,from both textual and speech input.
	
\end{abstract}
%
%
%\begin{figure}
%	\label{fig:cover}
%	\centering
%	\includegraphics[width=8cm,height=5cm]{cover.png}
%	\caption{Illustrating the core idea of learning a constrained embedding space mapping for text to image generation.}
%\end{figure}

\vspace*{-0.5cm}
\section{Introduction}
\label{sec:intro}

Humans are capable of using information from multiple sources or modalities to learn concepts that relate them together.   {Previous research} \cite{shams} has also shown that perceiving  {and relating} multiple modalities of input data is a key component for efficient learning. {Multi-modal data here, refers to  information from multiple sources. We use such multi-modal data for cognition in our every-day life.} However, in artificial learning systems, relating such modality independent concepts still remains largely an unsolved problem.

The primary reason for this is the difficulty to model such relationships in machines, since they do not generalize well to associate concepts with unseen objects from the training set. This failure to generalize novel concepts can be attributed to two main reasons : (1) Humans are constantly learning from multi-modal data since birth, which results in much better model learning and mapping between modalities compared to machine learning systems which typically only have access to a small finite subset of sample data. (2) The machine learning algorithms themselves are not yet effective for learning meaningful concepts in a generative manner from  training examples. Thus, multi-modal learning is an area of active research.

In this paper, we specifically target cross-modality concept learning, for the case of text and speech to images as depicted in figure 1. Some recent work have achieved success in this direction, i.e. to jointly learn generative models capable of generating one modality from another. For instance, Ngiam et. al. \cite{multimodal} proposed a deep learning framework using restricted Boltzmann machines \cite{hinton2002training} and deep belief networks \cite{hinton2006fast} to learn efficient features of audio and video modalities. They further illustrated that multi-modal learning results in better performance as compared to the unimodal case. In the case for learning to generate images from text modality, the recent work of Mansimov et. al.\cite{Salak} show that using attention-based models for generating images from text captions results in higher quality samples. Furthermore, it was claimed that this leads to better generalization towards previously novel captions. In other works, Reed et. al \cite{Scott} proposed deep convolutional generative adversarial networks which combined natural language and image embeddings in order to produce compelling synthetically generated images. Recently, there has also been some work in the field of cross-domain feature learning for images. Coupled generative models \cite{liu2016coupled}  generates pairs of images in two different domains by sharing weights for higher level feature extracting layers. Similarly other methods like Disco-GAN \cite{kim2017learning} and conditional VAEs \cite{kingma2014semi}, can learn to transfer style between images. However, prior works do not illustrate how different data modalities (like image, speech and natural language) can share similar features which can be used for conditional generation of different modalities. 

We rise to the challenge of this problem by jointly learning the distribution of multiple modalities of data using learned generative models of low-dimensional embeddings from high dimensional natural data. Our approach consists of first projecting the high dimensional data to a low-dimensional manifold or latent space and then separately, learn generative models for each such embedding space. In order to tie the two together we add an additional constraint on the learning objective to make  the two latent representations of each generative model be as close as possible. At inference time, the latent representations of one generative model can be used to proxy the other allowing a combined conditional generative model of multi-modal data. Using text, speech and image modalities, we show that our proposed method can successfully learn to generate images from modified MNIST datasets from text captions and speech snippets not seen during training. 

\vspace*{-0.2cm}

\begin{figure}[h]
	\label{fig:summary}
	\centering
	\includegraphics[width=7cm,height=4.5cm]{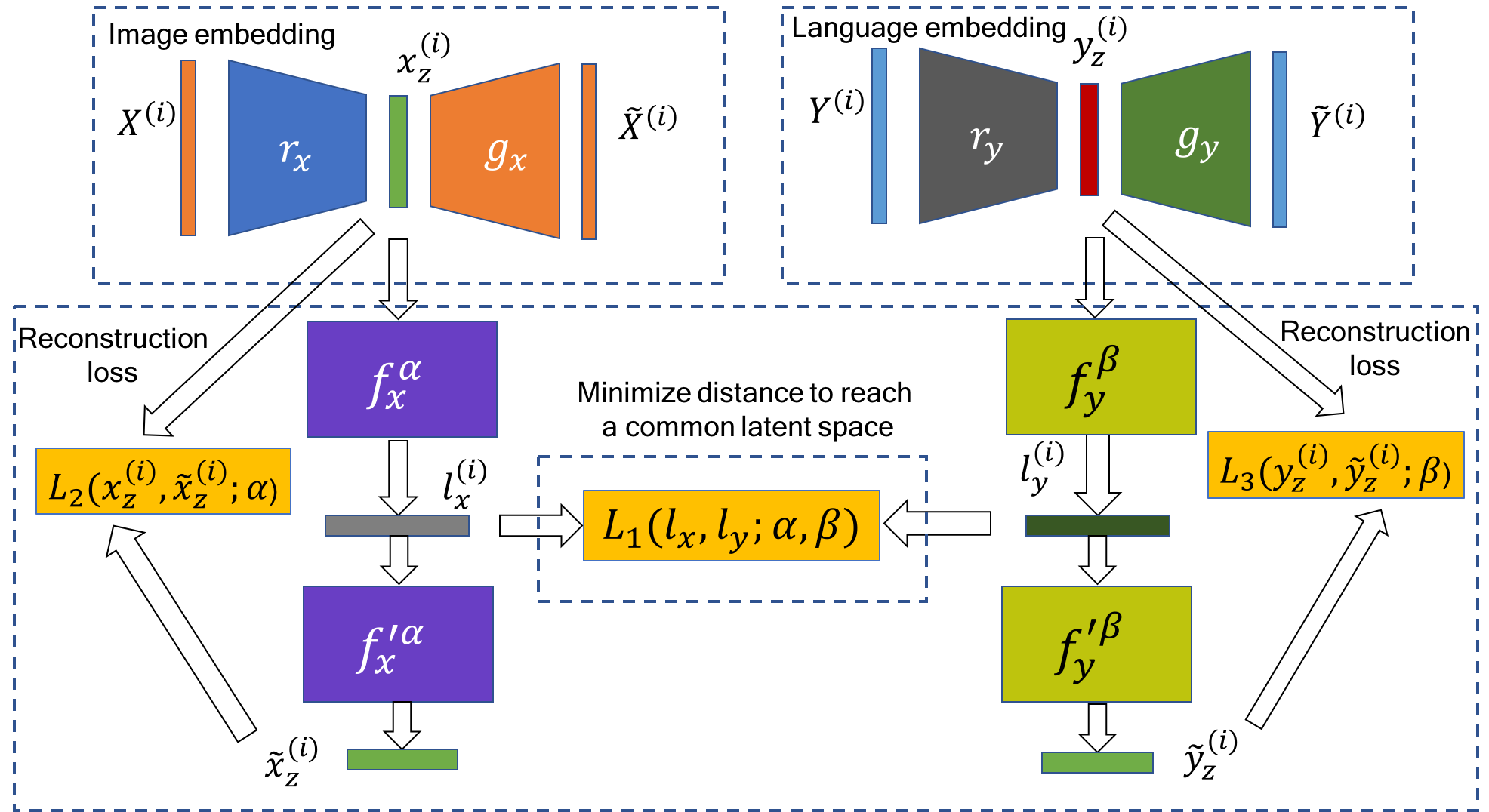}
	\caption{Illustration of the proposed method, which learns a mapping between embedding spaces by reducing the distance  $L_1(\bm{l_x},\bm{l_y};\bm{\alpha}, \bm{\beta})$ for  latent representations $\bm{l_x}$ and $\bm{l_y}$, while also ensuring that the inverse functions $f'_x, f'_y$ learns to generate the embeddings back by reducing the reconstruction errors $L_2(\bm{x_z},\bm{\widetilde{x}_z};\bm{\alpha})$ and $L_3(\bm{y_z},\bm{\widetilde{y}_z};\bm{\beta})$}
\end{figure}

\vspace*{-0.2cm}
\vspace*{-0.4cm}
\section{Problem Statement}
\label{sec:ps}
\vspace*{-0.2cm}
In this section, we formulate the problem of mapping between multiple modalities of data to learn semantic concepts. Consider two random variables $\textbf{X}$ and $\textbf{Y}$ representing instances of two input data modalities. Given the data, $ \mathbb{D}=\{\bm{x^i, y^i} \}$, ideally we would want to create a parametric joint distribution model $\log p_{\bm{\theta}}(\textbf{X},\textbf{Y}) = \sum_{i=1}^K \log{p_{\bm{\theta}}(\bm{x^i,y^i})}$ and find the optimal parameters $\bm{\theta ^{*}}$ as follows,

\vspace*{-0.5cm}
\begin{equation}
	\bm{\theta ^ *} = \argmin _{\bm{\theta}} \Big [-\sum_{i=1}^K \log{p_{\bm{\theta}}(\bm{x^i,y^i})} \Big],
\end{equation}
\vspace*{-0.5cm}

\noindent  where $K$ is the number of samples in the dataset. {Furthermore, learning the joint probability distribution enables us to perform conditional inference in order to map between the two data modalities $\textbf{X}$ and $\textbf{Y}$, i.e.}  $p_{\bm{\theta^*}}(\textbf{Y}|\textbf{X}=\bm{x}^j)$ and $p_{\bm{\theta^*}}(\textbf{X}|\textbf{Y}=\bm{y}^j)$.

\vspace*{-0.2cm}
\section{Proposed Method}
\label{sec:method}

\vspace*{-0.2cm}

\subsection{Learning the constrained embedding mapping}
\label{subsec:train}
\vspace*{-0.2cm}
The original high dimensional data points are first mapped to semantically meaningful low-dimensional manifold by deterministic or stochastic functions, $r_{X}  \colon \mathbb R^{N^x} \to \mathbb R^{d^x} $ and $r_Y \colon \mathbb R^{N^y} \to \mathbb R^{d^y}$, where $N^x, N^y$ are the dimensions of original high dimensional space and $d^x, d^y$ are the low dimensional embedding. Let the embeddings be represented as $\bm{x_z}, \bm{y_z}$ respectively. We also assume that the original high-dimensional data can be recovered from these embeddings by generative models. Such that, $g_{X} \colon \mathbb R^{d^x} \to \mathbb R^{N^x} $ and $g_Y \colon \mathbb R^{d^y} \to \mathbb R^{N^y}$.  We explicitly learn the mapping between them in a deterministic manner, while ensuring that the mapping can reconstruct or decode the original embeddings from the common latent space. We take inspiration from \cite{levine}, where the authors proposed a similar mapping to transfer learned skills between robots, and extend it to learn meaningful mappings among embedding spaces. 

Instead of modeling the original high dimensional joint distribution, $p_{\theta}(\textbf{X},\textbf{Y}) $, we model the joint distribution of the embedding space by learning a common latent variable($\bm{l}$) for both embeddings, such that $p_{\theta}(\bm{x_z},\bm{y_z}) = \int_l p_{\theta}(\bm{x_z},\bm{y_z}|\bm{l})p(\bm{l})d\bm{l}$. However, with a shared latent representation, both modalities of the data would be required to predict the latent space during inference, as $p(\bm{l}|\bm{x_z},\bm{y_z})$. To resolve this problem, we employ a proxy-variable trick, such that, we separately learn the generative models $p_{\alpha}(\bm{x_z})$ (with latent space $\bm{l_x}$) and $p_{\beta}(\bm{y_z})$ (with latent space $\bm{l_y}$) of the low-dimensional embeddings using an auto-encoder architecture parameterized by the network parameters $\bm{\alpha}$ and $\bm{\beta}$ respectively. We introduce an additional constraint that minimizes the distance between the latent representations $(\bm{l_x}, \bm{l_y})$ for each auto-encoder structure. As such, within this framework, reducing the original high dimensional data to low-dimensional embedding subspaces ensure two important purpose: (1) Since the embedding spaces are semantically meaningful, the intuition is that mapping between such subspaces lead to better generalization for novel data points, than using the original high dimensional data. (2) Learning a mapping on the embedding space allows us to exploit the generative capabilities of the separate models which can be trained from the plethora of unlabeled data. Furthermore, since we use a simple auto-encoder model, reduction in data dimension leads to better faster convergence with shallower networks.

\begin{figure}
	\label{fig:double}
	\centering
	\includegraphics[width=8.5cm,height=8.5cm]{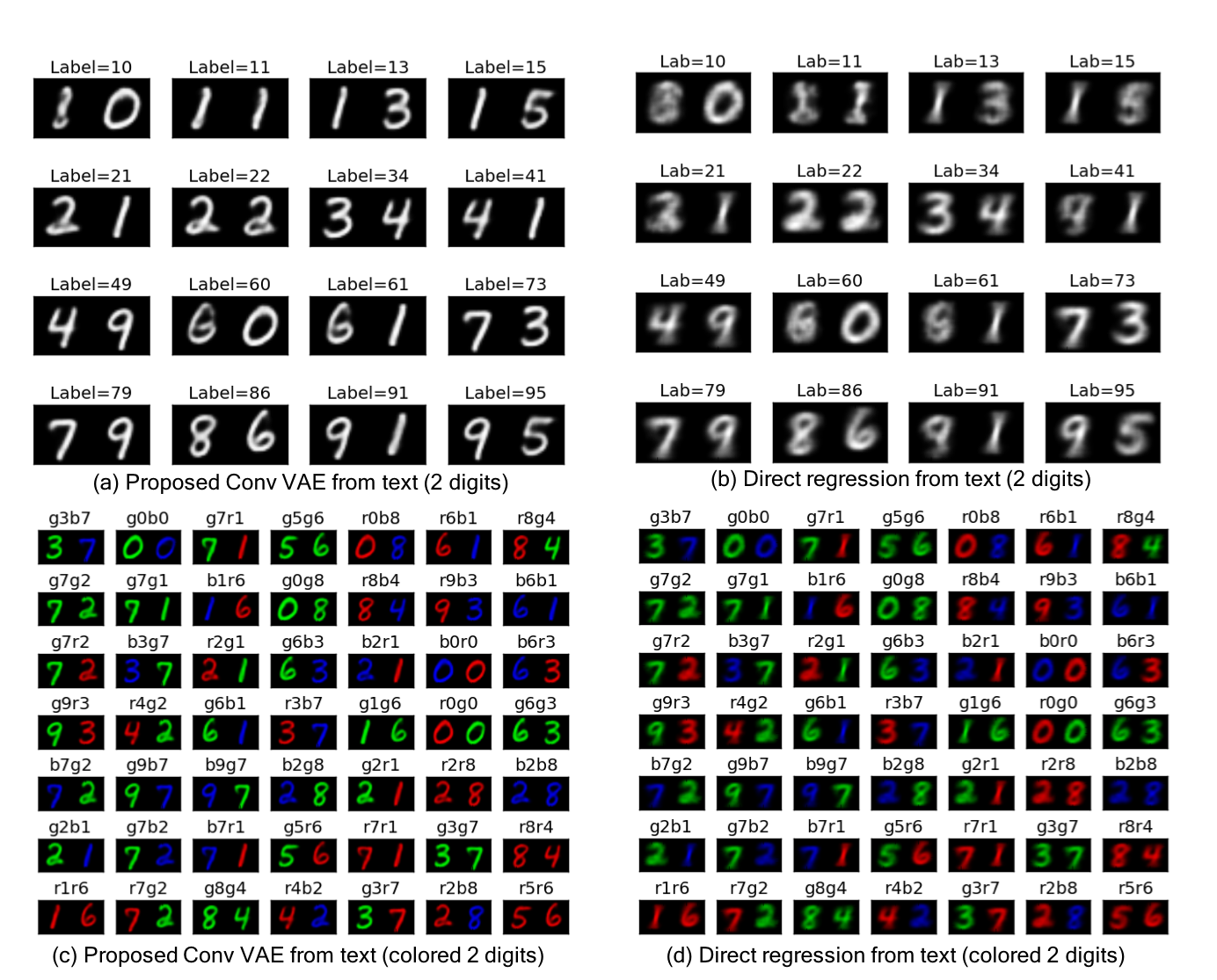}
	\caption{Generating images from text embeddings unseen during training for MNIST 2 digits and colored 2 digits.}
\end{figure}

%\begin{figure*}[h!]
%	\label{fig:col-double}
%	\centering
%	\includegraphics[width=18cm,height=6cm]{colored-double-mnist}
%	\caption{(a) t-SNE plot of image embeddings from test images (marked "x") and mapped word embeddings (marked black "o"), (b) Generating novel double digit combinations with color using convolutional VAE (c) The baseline method of word-embedding to image regression produces blurry mean images for color-digit combinations. (best seen in color, zoomed in)}
%\end{figure*}

The auto-encoders for image and text(or speech) embedding generation are parameterized by $\bm{\alpha}$ and $\bm{\beta}$ respectively. The intermediate latent representations are $\bm{l_x}=f_x(\bm{x_z};\bm{\alpha})$ and $\bm{l_y}=f_y(\bm{y_z};\bm{\beta})$ and  the reconstructed embeddings are computed by their inverse functions as, $\bm{\widetilde{x}_z}=f'_x(\bm{l_x};\bm{\alpha})$ and $\bm{\widetilde{y}_z}=f'_y(\bm{l_y};\bm{\beta})$. Reconstruction losses $L_2(\bm{x_z},\bm{\widetilde{x}_z};\bm{\alpha})$ and $L_3(\bm{y_z},\bm{\widetilde{y}_z};\bm{\beta})$ are also minimized in order to ensure that the respective embeddings can be generated back. Additionally, the two latent representations are constrained to be equal by minimizing the distance between them $L_1(\bm{l_x},\bm{l_y};\bm{\alpha}, \bm{\beta})$.  This facilitates multi-modal mapping during the inference step. We explain this in section \ref{subsec:mapping}. {Combining the three loss functions, the optimal parameters for the two auto-encoders are learned by minimizing the following objective function,}

\begin{equation}
	\label{eq:cost}
	\begin{aligned}
		\bm{\alpha ^  *,  \beta ^ *} = \argmin_{\bm{\alpha, \beta}}  L_1(\bm{l_x},\bm{l_y};\bm{\alpha}, \bm{\beta}) + L_2(\bm{x_z},\bm{\widetilde{x}_z};\bm{\alpha}) \\
		+ L_3(\bm{y_z},\bm{\widetilde{y}_z};\bm{\beta}) ,
	\end{aligned}
\end{equation}

In figure 1, we provide the outline of our training process. Once the networks are trained, we can rearrange each individual components in order to create a generative mapping model which is discussed in the following sub-section.

\begin{table*}[ht!]
	\centering
	\label{my-label}
	\begin{tabular}{|c|c|c|c|c|}
		\hline
		\multirow{2}{*}{} & \multicolumn{2}{c|}{\textbf{\begin{tabular}[c]{@{}c@{}}PSNR from \\ textual data (in dB)\end{tabular}}} & \multicolumn{2}{c|}{\textbf{\begin{tabular}[c]{@{}c@{}}PSNR from \\ speech data (in dB)\end{tabular}}} \\ \cline{2-5} 
		& Two digit & Colored two digit & Two digit & Colored two digit \\ \hline
		\textbf{Direct (conv)} & 14.73 & 19.12 & 15.26 & 19.24 \\ \hline
		\textbf{Proposed (conv-vae)} & \textbf{15.82} & \textbf{19.49} & \textbf{16.32} & 19.66 \\ \hline
		\textbf{Proposed (mlp-vae)} & 15.73 & 18.89 & 16.08 & 19.51 \\ \hline
		\textbf{Proposed (conv-ae)} & 14.97 & 18.90 & 15.26 & \textbf{19.67} \\ \hline
		\textbf{Proposed (mlp-ae)} & 14.89 & 18.39 & 15.22 & 19.05 \\ \hline
	\end{tabular}
	\caption{Quantitative comparison of the proposed method using different generative models compared to the baseline of directly predicting the images from text and speech embeddings by a convolutional network.}
\end{table*}

\subsection{Mapping between multi-modal data}
\label{subsec:mapping}

Since the networks were trained to respect the constraint of equality between the latent spaces, we can consider them to represent the same space during inference. However, even if the spaces $\bm{l_x}$ and $\bm{l_y}$ are equal, they are decoupled by design, which allows us to use them for generating specific modalities of data.

From the learned mapping, the original high-dimensional data can be recovered by generator functions, $ \widetilde{\textbf{X}}=g_X(\bm{\widetilde{x}_z})$ and $ \widetilde{\textbf{Y}}=g_Y({\bm{\widetilde{y}_z}})$. The mapping from language to image (image generation) can be performed as $ \widetilde{\textbf{X}}|\textbf{Y}=g_X(\bm{\widetilde{y}_z}|\bm{x_z})$ and image to language (image captioning) can be performed as $ \widetilde{\textbf{Y}}|\textbf{X}=g_Y(\bm{\widetilde{x}_z}|\bm{y_z})$.

\section{Experimental results}
\label{sec:expres}

Using the proposed conditional generative model, we generate images from corresponding text and speech representations. As an example, we use text "seven five" and also the corresponding speech representation, and independently learn to generate images of $75$ from each of the above two modalities.

For generating the embedding space from images, we evaluate our method using Multi-layered perceptron (MLP) based variational auto-encoders (mlp-vae) \cite{vae}, convolutional networks based variational auto-encoders (conv-vae), MLP based auto-encoders (mlp-ae) and Convolutional auto-encoders (conv-ae). For generating the word embeddings we use word2vec as proposed in \cite{word2vec} which was trained with wikipedia word corpus. For the speech signals, we use Mel-Frequency Cepstral Coefficients (MFCC) features as speech embeddings. For generating back text and speech from the embeddings, we use nearest neighbor to retrieve the closest data point to the query. In the image embedding case, we use a normalization function as the forward function, $f_X(\bm{a})=\frac{\bm{a}-\mu_x}{\sigma_x}$ and the corresponding un-normalization function as the inverse function, $f'_X(a)=a * \sigma_x + \mu_x$, where $\mu_x$ and $\sigma_x$ are the mean and standard deviation of image embedding in the training set. For the word and speech embeddings, we represent the non-linear mapping functions as neural networks. Specifically in our implementation, we use simple fully connected networks, containing 1 hidden layer in encoder and decoder units. L2 loss is used for all the components of the objective function. Adam optimizer is used with default parameters for minimizing the objective function in equation \ref{eq:cost}.

%\begin{table*}[]
%	\centering
%	%\caption{My caption}
%	\label{my-label}
%	\begin{tabular}{|c|c|c|c|c|}
%		\hline
%		\multirow{2}{*}{} & \multicolumn{2}{c|}{\textbf{\begin{tabular}[c]{@{}c@{}}PSNR from \\ textual data (in dB)\end{tabular}}} & \multicolumn{2}{c|}{\textbf{\begin{tabular}[c]{@{}c@{}}PSNR from \\ speech data (in dB)\end{tabular}}} \\ \cline{2-5} 
%		& Two digit & Colored two digit & Two digit & Colored two digit \\ \hline
%		\textbf{Direct (conv)} & 14.73 & 19.12 & 15.26 & - \\ \hline
%		\textbf{Proposed (conv-vae)} & \textbf{15.82} & \textbf{19.49} & \textbf{16.32} & \textbf{-} \\ \hline
%		\textbf{Proposed (mlp-vae)} & 15.73 & 18.89 & 16.08 & - \\ \hline
%		\textbf{Proposed (conv-ae)} & 14.97 & 18.90 & 15.26 & - \\ \hline
%		\textbf{Proposed (mlp-ae)} & 14.89 & 18.39 & 15.22 & - \\ \hline
%	\end{tabular}
%\caption{Quantitative comparison of the proposed method using different generative models compared to the baseline of directly predicting the images from text embeddings by a convolutional network. Showing the error between image embeddings predicted by image encoder and mapped image embeddings from text}
%\end{table*}

% Please add the following required packages to your document preamble:
% \usepackage{multirow}

Since the generated images will have some different attributes to the images in test dataset, it is difficult to directly compare them with arbitrary instances from the test set. Thus, we first find the image in test set ($\{\bm{x_{test}^i}\}_i$) that is closest (by euclidean distance measure) and find Peak Signal to Noise ratio (PSNR) of the generated image ($\bm{x_{pred}}$) as compared with this image, given as, 

\begin{equation}
	\label{eq:PSNR}
	\mbox{PSNR} = 20 \log (\frac{\mbox{maximum pixel intensity}}{\sqrt{\min_i ||x_{pred} - x^i_{test}||_2^2}})
\end{equation}

As a baseline method, we use direct mapping from word(or speech) embedding space to image pixel space using convolutional neural networks similar to the discriminator network in the DCGAN \cite{radford} architecture. This method of directly regressing the image from word embeddings does not involve generation from latent variables and thus it learns a mean image representations from each word embedding, which is demonstrated later in this section.

% Please add the following required packages to your document preamble:
% \usepackage{multirow}

%\begin{table}[]
%	\centering
%	\label{my-label}
%	\begin{tabular}{|c|c|c|}
%		\hline
%		\textbf{\begin{tabular}[c]{@{}c@{}}Generative \\ Methods\end{tabular}} & Two digit & Colored two digit \\ \hline
%		Proposed (conv-vae) & 0.17 $\pm$ 0.048 & 0.15 $\pm$ 0.089 \\ \hline
%		Proposed (mlp-vae) & 0.22 $\pm$ 0.04 & 0.15 $\pm$ 0.092 \\ \hline
%		Proposed (conv-ae) & 0.30 $\pm$ 0.11 & 0.36 $\pm$ 0.21 \\ \hline
%		Proposed (mlp-ae) & 0.64 $\pm$ 0.146 & 0.516 $\pm$ 0.26 \\ \hline
%	\end{tabular}
%	\caption{My caption}
%\end{table}

Experimental evaluation were performed to generate MNIST images from textual and speech embeddings. In all the four cases, we use a 100 dimensional image embedding of the images and corresponding word embeddings of dimension 13 for each word. speech embeddings of dimension 13 were used for each digits audio (sampling rate 24 KHz, each digit has 0.5 secs duration).

Double digit numbers were created by concatenating MNIST digit images horizontally. While training the image auto-encoder, we randomly remove some two-digit combinations and train on the remaining images. Thus the auto-encoder have explicitly never been trained to generate these images. For the word and speech embeddings, we simply concatenate the respective embeddings of each digit. While learning the mapping between image to word and speech embeddings, we hide the mapping between the unseen images mentioned above and learn mapping only on the remaining images-word combination. During testing, we give the word embeddings of those unseen 2-digit numbers and generate the corresponding images. 

Similar experiment with the addition of color attribute (to increase the complexity of the mapping) was also performed. Colors are red, green and blue only. Figure 2(c,d) and fig 4(c,d) in appendix) shows colored MNIST images.

%\begin{figure*}
%	\label{fig:mnist}
%	\centering
%	\includegraphics[width=15cm,height=8cm]{single-mnist}
%	\caption{(left) Generating the images from word embeddings of labels mentioned on top of each image. This verifies that the network successfully learns to map language to image using the proposed architecture. (right) t-SNE plot of the generated image embeddings from word embeddings. This illustrates that the mapping makes use of the semantic information is }
%\end{figure*}

\subsection{Qualitative Evaluation}

Figure 2 shows the generated double digit images from text combinations that were unseen during training. Mapping using convolution VAE(figure 2(a)) as image auto-encoder produces clear images for the word embeddings. MLP VAE (not shown in the image) produces images that are comparatively blurry compared to conv-vae case. For direct regression(figure 2(b)), the network learns to generate mean images of digits from the training dataset which can be seen for some digits like "one" and "five". It is to be noted that while direct regression learns to associate word embeddings with blurry mean images that it encountered during training, the proposed method finds mappings in the semantically meaningful image embedding space and subsequently generates clearer images, hence producing higher quality images. We see similar results for colored double mnist data-set in figure 2(c) and 2(d), where convolutional VAE produces better generalization for the novel color and digit combinations compared to the baseline direct regression method. Similar results are seen for image generation from speech in figure 4 as show in the supplementary materials.

\subsection{Quantitative Evaluation}

To quantitatively evaluate the performance of our algorithm, we take the mean of PSNR value of all double digits and all colored-double digit combination, computed by equation \ref{eq:PSNR}. Table 1 shows the PSNR values for both experiments using different image encoder-decoder methods and direct regression from text embeddings and speech features. The PSNR values show that convolutional VAEs produce the highest PSNR value for both experiments from text embeddings and double MNIST from speech features whereas convolutional auto-encoder produces best PSNR values for image generation from speech data colored MNIST double digits. Direct methods suffers from low PSNR because the blurry nature of the digits cause high divergence from the closest image in test set.

% For the colored MNIST case, direct regression produces second best PSNR after conv-vae, while other models suffer from low PSNR. To investigate the effectiveness of the mapping in the common latent space, we find the mean square error between projection from image and word embeddings onto the latent space. As evident from the results in Table 1, conv-vae shows minimum mean square error between the projections in the image embedding space, which explain the high PSNR of the reconstructed images. Additionally, it is evident that there is an inverse proportionality between the mean errors in latent space and the corresponding image generation quality. 

\section{Conclusion}
\label{sec:conclusion}
We propose a multi-modal mapping model that can generate images even from unseen captions. The core of the proposed algorithm is to explicitly learn separate generative models for low-dimensional embeddings of multi-modal data. Thereby, enforcing the key equality constraint between latent representations of the data enables two-way generation of information. We showcase the validity of our proposed model by performing experimental evaluation on generating two-digit MNIST images and colored two digit numbers from word embeddings and speech features not seen during training. In the future, we hope to extend our method for generating images from complex captions using higher dimensional natural images.

\bibliographystyle{icml2017}
\bibliography{reference}
\clearpage
\begin{center}
	\huge{\textbf{Supplementary \\ material}}
\end{center}
\begin{appendices}
	
	\section{Network description}
	\label{whatever1}
	In this section we provide details of the network architecture to ensure reproducibility of the paper. For all networks, adam optimizer is used with default values in keras neural network library. For the image auto-encoders binary cross-entropy was used as reconstructions loss. For mapping embeddings to latent space, mean squared error was used as loss function.

%\title{Appendix}
\subsection{Image auto-encoder}
	
For both the case of double MNIST and colored double MNIST, we use same configuration for the network parameters.
\begin{itemize}
\item \textbf{Convoutional Variational Auto-encoder}:
For the encoder, we used 2D convolution layers with $8$ filters of size $5 \times 5$ each with $2 \times 2$ max-pooling, followed by dense layers of size 256 and two dense layers of 100 dimension for mean and standard deviation of the image embedding space. For the decoder, we used a dense layer of size $3136$, followed by 2D convolution layer with $8$ filters of size $5 \times 5$ each with $2 \times 2$ up-sampling and 2D convolution layer of size $5 \times 5$ with single channel output to produce the image back. For all the layers, non-linearity of ReLU was used except the last convolution layer in the decoder which used a sigmoid non-linearity.
	
\item \textbf{MLP Variational Auto-encoder}:
For the encoder, we used a hidden dense layers of size 256 and two dense layers of 100 dimension for mean and standard deviation of the image embedding space. For the decoder, we used a hidden dense layers of size 256 and a dense layers of 1568 dimension to produce the image back. For all the layers, non-linearity of ReLU was used except the last dense layer in the decoder which used a sigmoid non-linearity.
	
\item \textbf{Convoutional Auto-encoder}:
For the encoder, we used 2D convolution layers with $16$ filters of size $3 \times 3$ each with $2 \times 2$ max-pooling, followed by another 2D convolution layers with $8$ filters of size $3 \times 3$ and $2 \times 2$ max-pooling, followed by dense layers of 100 dimension for the image embedding space. For the decoder, we used a dense layer of size $784$, followed by 2D convolution layer with $8$ filters of size $3 \times 3$ each with $2 \times 2$ up-sampling and another 2D convolution layer with $16$ filters of size $3 \times 3$ each with $2 \times 2$ up-sampling. Finally a 2D convolution layer of size $5 \times 5$ with single channel output si used to produce the image back. For all the layers, non-linearity of ReLU was used except the last convolution layer in the decoder which used a sigmoid non-linearity.
		
\item \textbf{MLP Auto-encoder}:
For the encoder, we used a hidden dense layers of size 256 and a dense layers of 100 dimension for the image embedding space. For the decoder, we used a hidden dense layers of size 256 and a dense layers of 1568 dimension to produce the image back. For all the layers, non-linearity of ReLU was used except the last dense layer in the decoder which used a sigmoid non-linearity.
\end{itemize}
	
\begin{figure}[h]
\label{fig:appendix-fig}
\centering
\includegraphics[width=8cm,height=8cm]{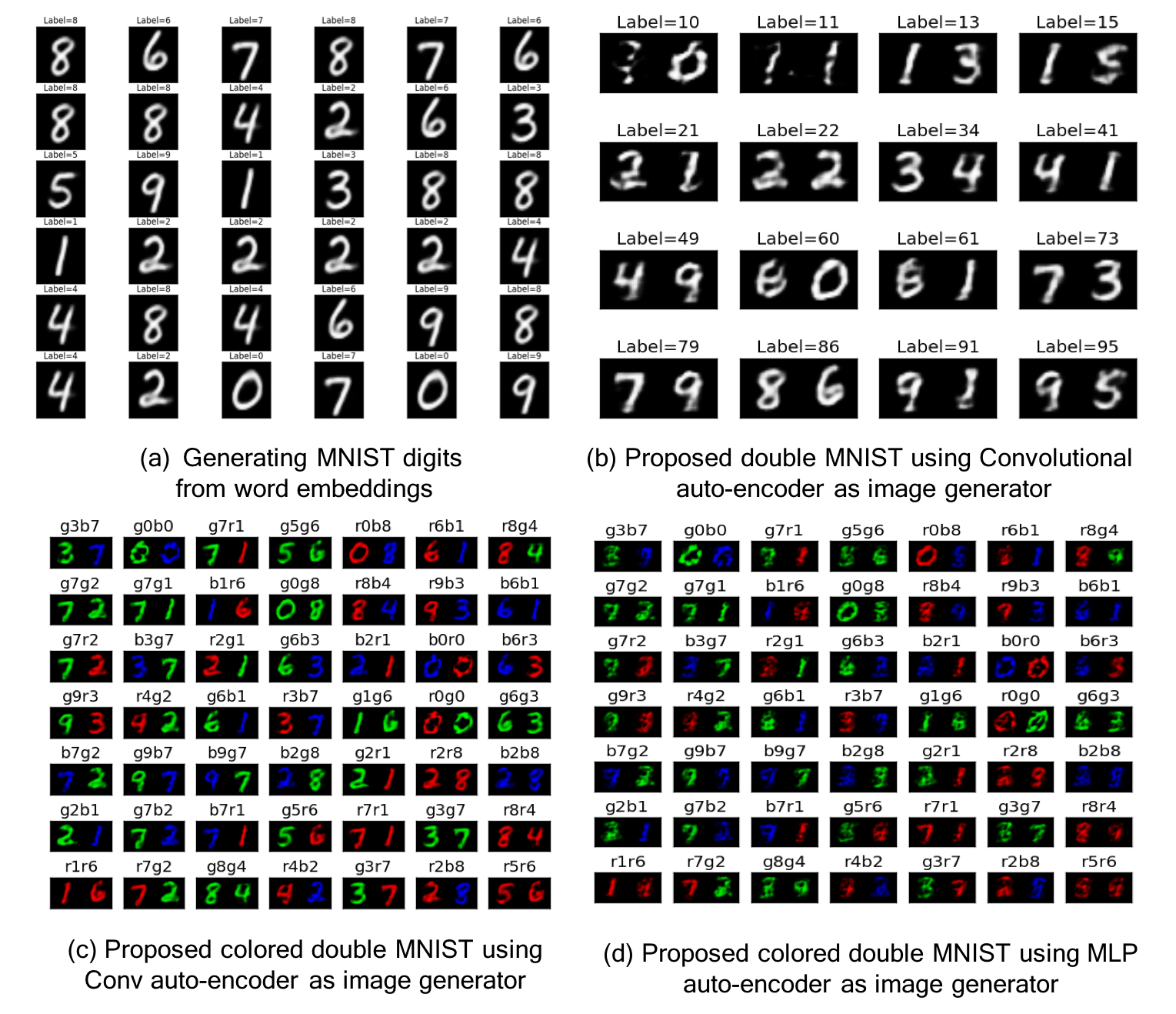}
\caption{Providing additional results for reconstructed MNIST digits with single and double + colored digits attribute.}
\end{figure}
\subsection{Mapping module network architecture}

For image embedding, we use a normalization function as the forward function, $f_X(\bm{a})=\frac{\bm{a}-\mu_x}{\sigma_x}$ and the corresponding un-normalization function as the inverse function, $f'_X(a)=a * \sigma_x + \mu_x$, where $\mu_x$ and $\sigma_x$ are the mean and standard deviation of image embedding in the training set. In other words, we force the shared latent space to be the normalized image embedding space in our implementation. For the word embeddings, we represent the non-linear mapping functions as neural networks. The encoder from the word-embedding to the common latent space contains a hidden dense layers of size 256 and a dense layers of 100 dimension for the common latent space.
	
\section{Additional results}
\label{whatever2}
We present additional results for image generation from text embeddings, in figure 3, using various types of auto-encoders as image generators. Qualitative inspection reveals that convolutional VAE produces the clearest and best-looking results, which is in accordance to the PSNR values in Table 1. 
For the image generation from speech embeddings (MFCC features), please refer to figure 4. Qualitative inspection reveals that proposed method produce clear and good-looking results compared to direct regression baseline, which is in accordance to the PSNR values in Table 1.  
	
\begin{figure}[h]
\label{fig:audio}
\centering
\includegraphics[width=9cm,height=9cm]{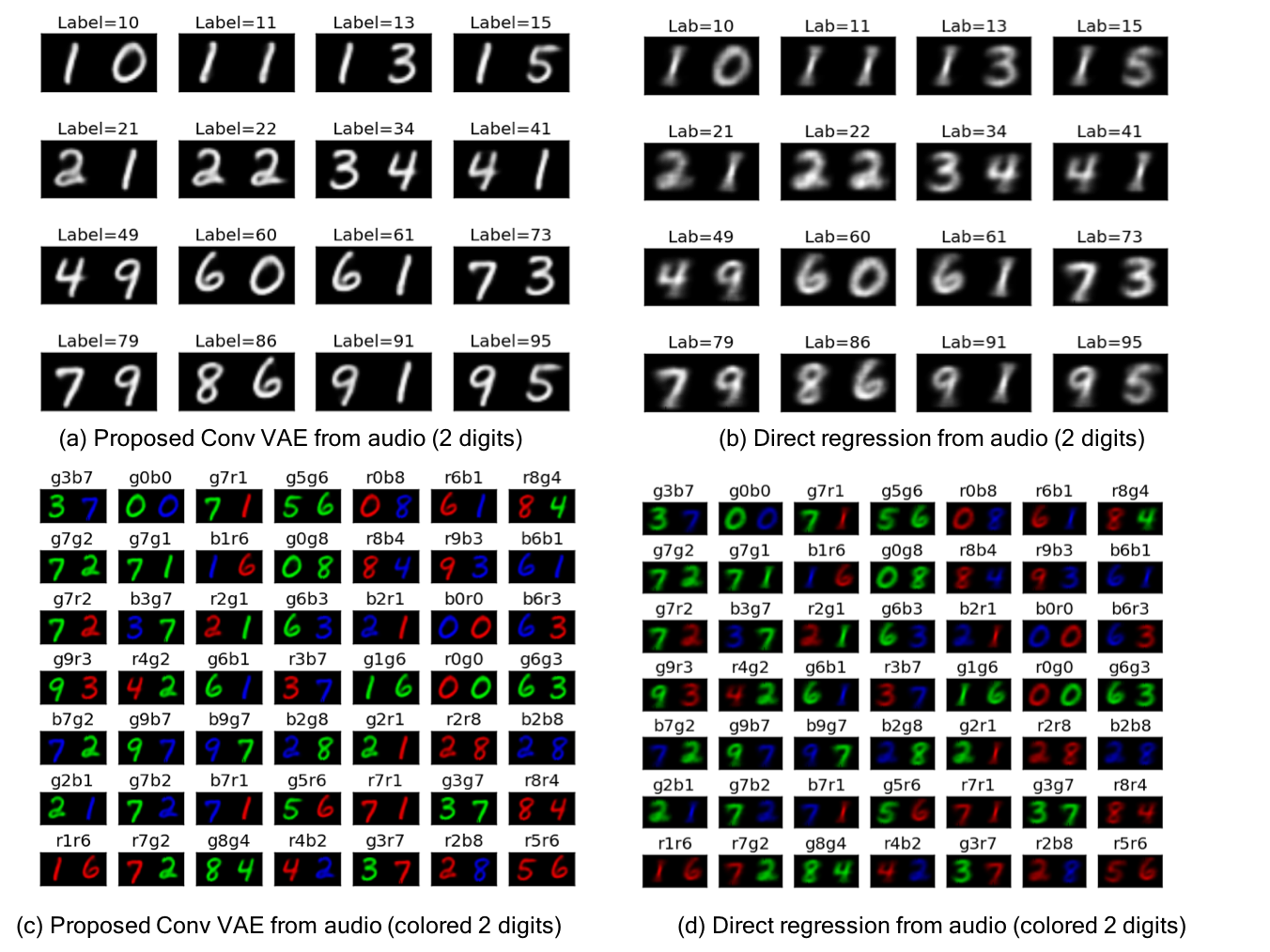}
\caption{Generating images from speech sequences unseen during training for MNIST 2 digits and colored 2 digits.}
\end{figure}
	
\section{Experimental details}
Experimental evaluation were performed on MNIST dataset for four cases for both speech and text data. it is to be noted that, for both speech and text data, the embeddings are fixed for each class. Thus each class in the double digits MNIST case (total classes : 100) and colored double digit MNIST (total classes : 900) have singleton speech signal and text embedding.
	
\begin{itemize}
\item \textbf{Double MNIST digits } :
In this experiment, we attempt to learn the concept of double digits by testing our algorithm to generate novel images of double digit combinations from text embeddings. Double digit numbers (total 100 classes) are created by concatenating MNIST digit images horizontally. For each such 2-digit combination, 1000 images are generated. During training of the image auto-encoder, we randomly remove set of sixteen such two-digit combinations and train on the remaining 84000 images. Thus the image embeddings have explicitly never been trained to create these images. For the text and speech embeddings, we simply concatenate the embeddings of each digit. While learning the mapping between image and text or speech embeddings, we hide the mapping between those 16000 images and learn mapping only on the 84000 images-(text, speech) combination. During testing, we give the text and speech embeddings of those sixteen 2-digit numbers and generate the corresponding images. 
		
\item \textbf{Colored double MNIST digits} :
This experiment is similar to above experiment with the addition of color attribute to increase the complexity of the mapping. Colors are red, green and blue only. For each such 2-digit combination, 4000 images are generated by randomly juxtaposing images of digits with random color. During training of image auto-encoder, we randomly remove set of sixteen such colored two-digit combinations and train on the remaining 336000 images. For the text and speech embeddings, we simply concatenate the embeddings of each word of the digit (for example, red 5 blue 1 is concat(word2vec("red","five","blue","one"))).
		
While learning the mapping between image and text or speech embeddings, we hide the mapping between those 64000 images and learn mapping only with the other 336000 images. During testing, we give the text and speech embeddings of those sixteen two-digit numbers with 9 color combinations (3 $\times$ 3) and generate the corresponding images.
		
\end{itemize}	
\end{appendices}
\end{document}